\documentclass[transmag]{IEEEtran}
\usepackage{latexsym}
\usepackage{graphicx}
\usepackage{amsfonts,amssymb,amsmath}
\usepackage{hyperref}
\usepackage{hhline}
\usepackage{array}
\usepackage{booktabs}
\usepackage{subfig}
\usepackage{color}

\newcommand{\bom}[1]{\boldsymbol{#1}}
\newcommand{\bo}[1]{\mathbf{#1}}

\newcommand{\tr}{\mathrm{tr}}     
\DeclareMathOperator*{\argmin}{\arg \min}
\newcommand{\PDHm}[1]{\mathbb{S}_{++}^{#1 \times #1}}

\newcommand{\beb}{\bo b}
\newcommand{\m}{\boldsymbol{\mu}}
\newcommand{\muh}{\hat{\m}}  
\newcommand{\Sig}{\bom \Sigma} 
\newcommand{\B}{\bo B}  
\renewcommand{\S}{\bo S}  

\newcommand{\Bh}{\hat{\mathcal B}}  
\newcommand{\bebh}{\hat{\beb}}  

\newcommand{\ndim}{n}             
\newcommand{\pdim}{p}             

\renewcommand{\b}{\bo b}
\newcommand{\x}{\bo x}    

\newcommand{\gt}{\tilde{g}}  

\newcommand{\X}{\bo X}   
\newcommand{\Sigt}{\hat{\boldsymbol{\Sigma}}} 
\newcommand{\Sigit}{\hat{\boldsymbol{\Sigma}}^{-1}}  
\newcommand{\R}{\mathbb{R}}    

\newcommand{\beq}{\begin{equation}}
\newcommand{\eeq}{\end{equation}}
\newcommand{\bmat}{\begin{pmatrix}}
\newcommand{\emat}{\end{pmatrix}}

\def\BibTeX{{\rm B\kern-.05em{\sc i\kern-.025em b}\kern-.08em T\kern-.1667em\lower.7ex\hbox{E}\kern-.125emX}}

\graphicspath{{./}{./pics/} }

\begin{document}

\title{A Compressive Classification Framework for High-Dimensional Data}

\author{Muhammad Naveed Tabassum \IEEEmembership{Member, IEEE} and Esa Ollila \IEEEmembership{Member, IEEE} 
\thanks{The research was supported by the Academy of Finland grant no. 298118 which is gratefully acknowledged}
\thanks{M.N. Tabassum and E. Ollila are with Aalto University, Dept. of Signal Processing and Acoustics, P.O. Box 15400, FI-00076 Aalto, Finland (e-mail: \{muhammad.tabassum, esa.ollila\}@aalto.fi)}}

\IEEEtitleabstractindextext{\begin{abstract}
We propose a compressive classification framework for settings where the data dimensionality is significantly larger than the sample size.  The proposed method, referred to as compressive regularized discriminant analysis (CRDA), is  based on linear discriminant analysis and has the ability to select significant features by using joint-sparsity promoting hard thresholding in the discriminant rule. Since the number of features is larger than  the sample size, the method also uses state-of-the-art regularized sample covariance matrix estimators.  Several  analysis examples on real data sets, including  image, speech signal and gene expression data illustrate the promising improvements offered by the proposed CRDA classifier in practise. Overall, the proposed method gives fewer misclassification errors than its competitors, while at the same time achieving accurate feature selection results.  The open-source R package and MATLAB toolbox of the proposed method (named compressiveRDA) is freely available.
\end{abstract}

}

\maketitle

\section{Introduction}\label{sec:intro}

High-dimensional (HD) classification is at the core of numerous contemporary statistical studies. Commonly large amounts of information may be collected on each individual sample point, yet the number of sample points themselves may remain relatively small.  Typical examples of such cases are  gene expression and protein mass spectrometry data,  and other areas of computational biology.  Regularization and shrinkage are commonly used tools in many applications such as regression or classification to overcome significant statistical challenges posed particularly due to the high dimension low-sample-size (HDLSS) data settings in which the number of features, $\pdim$, is often several magnitudes larger than the sample size, $\ndim$ (i.e., $\pdim \gg \ndim$).  In this paper, we propose a novel classification framework for such HD data. 

In many applications, the ideal classifier of an HD input vector should have three characteristics: (i) the proportion of correctly classified observations is high, (ii) the method is computationally feasible, allowing the training of the classifier in real-time with a personal computer, and (iii) the classifier selects only the significant features, thus yielding interpretable models. The last property is important for example in gene expression studies, where it is important to identify which genes influence the trait (e.g., disease). This is obviously important for the biological understanding of the disease process but can help in the development of clinical tests for early diagnosis \cite{tibshirani2003NSC}.  We propose a HD classification method that takes into account all of the aforementioned three key aspects.

Some of the standard off-the-shelf classification methods either lack feature elimination or require that it is done beforehand. For example, the basic support vector machine (SVM) classifier uses all features. Some other classifiers on the other hand have higher computational cost, e.g., Logit-ASPLS approach (with R-package: plsgenomics) \cite{logitASPLS2018}  depends on three hyperparameters. This increases the computational complexity of the method, especially when  finer grids of hyperparameters are used and tuned using cross-validation (CV). In this paper,  we propose a classification framework referred to as {\sl compressive regularized discriminant analysis (CRDA)} method, which is computationally efficient, performs feature selection during the learning process and includes the  shrunken centroids (SC)RDA method \cite{guo2006regularized} as a special case. Our simulation studies and real data analysis examples illustrate that CRDA often achieves superior classification performance compared to off-the-shelf classification approaches in the HDLSS settings. 
 
Many classification techniques assign a $\pdim$-dimensional observation $\x$ to one of the $G$ classes (groups or populations) based on the following rule:  
\beq\label{eq:ldarule} 
\x  \in \texttt{group}\, \big[\, \gt = \arg \max_{g} \: d_{g}(\x) \,\big],
\eeq
where $d_g(\x)$ is called the {\sl discriminant function} for population $g \in \{1,\ldots, G\}$. In  linear discriminant analysis (LDA),  $d_g(\x)$ is a linear function of $\x$,  $d_g(\x)= \x^\top \beb_g + c_g$, for some constant $c_g \in \R$ and vector $\beb_g \in \R^\pdim$, $g=1,\ldots,G$. The vector $\beb_g=\beb_g(\Sig)$ depends on the unknown covariance matrix $\Sig$ of the populations (via its inverse matrix) which is commonly estimated using the pooled sample covariance matrix (SCM). In the HD settings, the SCM is no-longer invertible, and  regularized SCM (RSCM) $\hat \Sig$ is used  for constructing an estimated discriminant function $\hat d_g(\x)$.  Such approaches are commonly referred to as regularized LDA methods, which we refer shortly as RDA. See e.g.,  \cite{guo2006regularized,tibshirani2003NSC,witten2011plda, sharma2014feature, neto2016regularized}.

Next note that if the $i$-th entry of $\beb_g$ is zero, then the $i$-th feature does not contribute in the classification of $g$-th population. To eliminate unnecessary features, one can shrink $\beb_g$ using element-wise soft-thresholding as in the shrunken centroids SCRDA method \cite{guo2006regularized}. These methods are often difficult to tune because the shrinkage threshold parameter is the same across all groups, but different populations would often benefit from different shrinkage intensity. Consequently, they tend to yield higher false-positive rates (FPRs).

Element-wise shrinkage does not achieve simultaneous feature elimination because the eliminated feature from group $i$ may still affect the discriminant function of  group $j$. 
We propose CRDA classification framework which generalizes the SCRDA method by allowing more general sparsifying transformation functions to be used in the linear discriminant rule.  In particular, we promote the usage of {\it joint-sparsity}  via $\ell_{q,1}$-norm based hard-thresholding.  Another advantage is of having a shrinkage parameter that is much easier to tune and interpret: namely, the {\sl joint-sparsity level} $K \in \{ 1, 2,\ldots, \pdim\}$ instead of a shrinkage threshold $\Delta \in [0, \infty)$ used in the soft-thresholding function of SCRDA.  We also promote the usage of two recent state-of-the-art regularized covariance matrix estimators proposed in \cite{ollila2019optimal} and \cite{yu2017high,tyler2019shrinking}, respectively. The first estimator, called the Ell-RSCM estimator \cite{ollila2019optimal}, is designed to attain the minimum mean squared error and is fast to compute due to automatic data-adaptive tuning of the involved regularization parameter.  The second estimator is a penalized SCM (PSCM)  proposed by \cite{yu2017high}, which we refer to as Rie-PSCM, that minimizes a penalized negative Gaussian likelihood function with a Riemannian distance penalty.  Although a closed-form solution of the optimization problem is not possible to find, the estimator can be computed using a relatively fast iterative Newton-Raphson algorithm developed in \cite{tyler2019shrinking}.  

We test the effectiveness of the proposed CRDA methods using partially synthetic genomic data as well as on nine real data sets and compare the obtained results against seven state-of-the-art HD classifiers that also perform feature selection. 
The results illustrate that overall, the proposed CRDA method gives fewer misclassification errors than its competitors while achieving  at the same time accurate feature selection with minimal computational efforts. Preliminary results of CRDA approach has appeared in \cite{crda2018}. 

The paper is structured as follows.  \autoref{sec:RLDA} describes the regularized LDA procedure and the regularized covariance matrix estimators that will be used by the CRDA classifiers.  In \autoref{sec:CRDA}, CRDA is introduced as a compressive version of regularized LDA performing feature  selection  simultaneously with classification.  Classification results for 3 different types (image, speech, genomics) of real data sets are also presented.  \autoref{sec:psd} illustrates the promising performance of CRDA in feature selection for partially synthetic data set while \autoref{sec:results} provides analysis results  for six real gene expression  data sets against a large number of benchmark classifiers. \autoref{sec:concl} provides discussion and concludes the paper.  MATLAB  and R toolboxes  are available at {\tt \url{http://users.spa.aalto.fi/esollila/crda}} and {\tt \url{https://github.com/mntabassm/compressiveRDA}}). 

{\it Notations:} We use $\| \cdot \|_{\mathrm{F}}$ to denote the Frobenius (matrix) norm, defined as 
  $\| \bo A \|_{\mathrm{F}}^2  = \tr(\bo A^\top \bo A)$ for all matrices $\bo A$.  An $\ell_q$-norm is defined as
\[
\| \bo a \|_q = \begin{cases}  \big( |a_1|^q+ \ldots + |a_n|^q\big)^{1/q}, & \mbox{for $q \in [1, \infty)$} \\ 
 \max \big( | a_i | \, : \, i=1,\ldots, n \big), &  \mbox{for $q=\infty$}  \end{cases}
 \]
for any vector $\bo a \in \R^n$.  The soft-thresholding (ST)-operator is defined as 
\[
\mathcal S_\Delta(b)=  \mathrm{sign}(b)(|b| - \Delta)_+ , 
\]
where  $(x)_+=\max(x,0)$  for $x \in \R$ and $\Delta>0$ is a threshold constant.  Finally,  $\mathrm{I}(\cdot)$ denotes the indicator function which takes value 1 when true and value 0 otherwise. $\mathbb{E}[\cdot]$ denotes expectation. 

\section{Regularized LDA} \label{sec:RLDA}

We are given a $\pdim$-variate random vector $\x$ which we need to classify into one of $G$ groups or populations. In LDA, one assumes that the class populations are $\pdim$-variate multivariate normal (MVN) with a common positive definite symmetric covariance matrix $\Sig$ over each class but having distinct class mean vectors $\boldsymbol{\mu}_g \in\mathbb{R}^{\pdim}$, $g=1,\ldots, G$.  The problem is then to classify  $\x$ to one of the MVN populations, $\mathcal N_p( \boldsymbol{\mu}_g, \Sig)$, $g=1,\ldots, G$. Sometimes prior knowledge is available on proportions of each population and we denote by $p_g$ the  prior probabilities of the classes ($\sum_{g=1}^G p_g = 1$). LDA uses the rule \eqref{eq:ldarule} with discriminant function
\[
d_g(\x)=   \x^\top \beb_g - \frac 1 2    \boldsymbol{\mu}_g ^\top  \beb_g   + \ln p_g,
\]
where $\beb_g = \Sig^{-1} \boldsymbol{\mu}_g$ for $g=1,\ldots, G$.  

The unknown parameters in LDA are the class mean vectors $\{\boldsymbol{\mu}_g\}_{g=1}^G$ and the covariance matrix $\Sig$. These are estimated from the {\sl training data set}  $\X=(\x_1 \, \cdots \, \x_n)$ which consists of $n_g$ observations from each class, $g=1,\ldots, G$. Let $y_i$ denote the class label associated with the  $i$-th observation, so $y_i \in \{ 1, \ldots, G\}$.  Then $ n_g = \sum_{i=1}^n \mathrm{I}(y_i=g)$ is the number of observations belonging to $g$-th population,  and we denote by  $\pi_g=n_g/n$  the relative sample proportions. We assume observations in the training data set $\X$ are centered by the sample mean vectors of the classes, 
\vspace{-0.2ex}
\beq \label{eq:muh}
\muh_g  = \bar{\x}_g = \frac{1}{n_g}\,\sum_{ y_i=g} \x_i.  
\eeq

Since the data matrix $\X$ is  centered group-wise,  the pooled (over groups) SCM can be written simply as 
\beq \label{eq:pooledSCM}
\S =  \frac{1}{n} \,\X \X^\top.
\eeq 
In practice, an observation $\x$ is classified using an estimated discriminant function, 
\beq\label{eq:edf} 
\hat d_g(\x) =  \x^\top \bebh_g - \frac 1 2  \muh_g^\top \bebh_g     + \ln \pi_g, 
\eeq
where $\bebh_g = \hat{\boldsymbol{\Sigma}}^{-1} \muh_g$, $g=1,\ldots, G$ and $\hat{\boldsymbol{\Sigma}}$ is an estimator of  $\Sig$. 
Note that in \eqref{eq:edf}  the prior probabilities $p_g$-s are  replaced by their estimates, $\pi_g$-s.  
Commonly, the pooled SCM $\S$ in \eqref{eq:pooledSCM} is used as an estimator $\hat{\boldsymbol{\Sigma}}$ of the covariance matrix $\Sig$. 

Since  $p \gg n$, the pooled SCM is no longer invertible.   To avoid the singularity of the estimated covariance matrix, a commonly used approach  ({\it cf.} \cite{ledoit2004well}) is to use  a \emph{regularized SCM (RSCM)} that shrinks the estimator towards a target matrix.  
One such method is reviewed in \autoref{sec:ell2rscm}. 
An alternative approach is to use penalized likelihood approach wherein a penalty term is added to the likelihood function. The idea is to penalize covariance matrices that are far from the desired target matrix.  One prominent estimator of this type is reviewed in \autoref{sec:riepscm}.  There are many other covariance matrix estimators for high-dimensional data, including Bayes estimators \cite{coluccia2015regularized}, 
structured covariance matrix estimators \cite{meriaux2019robust,wiesel2015structured}, 
 robust shrinkage estimators \cite{ollila2014regularized,pascal2014generalized,sun2014regularized,chen2010shrinkage},  etc, but  
 most of these are not applicable in HDLSS setting due to incurred high computational cost.

\subsection{Regularized SCM estimators} \label{sec:ell2rscm}

A commonly used RSCM estimator (see e.g., \cite{ledoit2004well,ollila2017optimal,ollila2019optimal}) is of the form:
\beq \label{eq:rscm} 
\Sigt(\alpha) = \alpha \,\S + (1-   \alpha) \,  [\tr(\S)/\pdim] \, \bo I ,
\eeq 
where  $ \alpha \in [0,1)$ denotes the shrinkage (or regularization) parameter and $\bo I $ denotes an identity matrix of appropriate size. 
We use the automated method for choosing the tuning parameter $\alpha$ proposed in \cite{ollila2019optimal}  and the respective estimators,  
called Ell1-RSCM and Ell2-RSCM. These estimators use $\hat \alpha$ that is a consistent estimator of $\alpha_o$  minimizing the mean squared error,  
\beq \label{eq:al_o}
\alpha_o =  \underset{\alpha \in [0,1)}{\arg \min}  \  \mathbb{E} \big[ \| \Sigt(\alpha) - \Sig \|_{\mathrm{F}}^2 \big]. 
\eeq 
See Appendix~\ref{app:alpha_est} for details. 
    
Note that SCRDA  \cite{guo2006regularized}  uses  a different  estimator, 
 $\Sigt =   \alpha \S + (1- \alpha) \bo I$, where the regularization parameter $ \alpha$ is  chosen using cross-validation (CV).  The benefit of Ell1-RSCM and Ell2-RSCM estimators are that they avoid computationally intensive CV scheme. In the HD settings, another computational burden is related to inverting the matrix $\Sigt$ in \eqref{eq:rscm}. The matrix inversion can be done efficiently using the SVD-trick detailed in Appendix~\ref{app:compute_siginv}.

\subsection{Penalized SCM estimator using Riemannian distance}  \label{sec:riepscm}

Assume that $\{ \x_i\}_{i=1}^\ndim$  is a random sample from a MVN distribution. Then the maximum-likelihood estimator of $\bom \Sigma$ is found by minimizing the $-(2/\ndim) \times$ Gaussian log-likelihood function,  
\[
L(\bom \Sigma; \bo S) =  \tr \left( \bom{\Sigma}^{-1}  \bo S \right) + \log   | \bom{\Sigma} | 
\]
over the set of positive definite symmetric $\pdim \times \pdim$ matrices, denoted by $\PDHm{\pdim}$.  
When $n > p$ and the SCM $\bo S$ is non-singular,   the minimum is attained at $\hat{\bom \Sigma}= \bo S$.  
Since  $p \gg n$ and $\bo S$ is singular,  it is common to add a penalty  function $\Pi(\bom \Sigma)$  to the likelihood function  to enforce a certain structure to the covariance matrix and to guarantee that the objective function has a unique (positive definite matrix) solution. One such penalized SCM (PSCM) proposed in \cite{yu2017high} uses a Riemannian distance of $\bom \Sigma$ from  a scaled identity matrix,  $ \Pi(\bom \Sigma; m)= \| \log(\bom \Sigma) -  \log(m) \bo I \|_{\mathrm F}^2$, where $m>0$ is  a shrinkage constant. One then solves the penalized likelihood function: 
\begin{eqnarray}\label{eq:precision-mat}
  \hat{\bom{\Sigma}} = \argmin_{\bom{\Sigma}\in \PDHm{\pdim}}  \,  L(\bom \Sigma; \bo S) + \eta \| \log(\bom \Sigma) -  \log (m) \bo I \|_{\mathrm F}^2 , 
\end{eqnarray}
where $\eta>0$ is the penalty parameter.  As in \cite{tyler2019shrinking},  we choose $m$ as the mean of the eigenvalues ($d_1 \geq  \cdots \geq  d_{p} \geq 0 $) of the SCM, i.e., 
\[
m= \frac{\tr(\bo S)}{p} = \frac 1 p \sum_{i=1}^p d_i .
\]
The eigenvalues of  $\hat{\bom \Sigma}$ are then shrunked towards $m$, and $\hat{\bom \Sigma} \to m \bo I$ in the limit when $\eta \to \infty$  \cite{yu2017high}.  
The solution to \eqref{eq:precision-mat}, which we refer to as  \emph{Rie-PSCM estimator}, does not have a closed-form solution, but a relatively efficient Newton-Raphson algorithm \cite{tyler2019shrinking}   can be used to compute the solution.   We choose the associated penalty parameter $\eta$ using the cross-validation procedure proposed in \cite{tyler2019shrinking}.

\section{Compressive RDA} \label{sec:CRDA}

In order to explain the proposed compressive RDA (CRDA) approach, we first write the discriminant rule in \eqref{eq:edf} in  vector form  as 
\begin{align} 
\hat{\bo d}(\x) &= ( \hat d_1(\x), \ldots, \hat d_G(\x) )  \notag  \\
&=  \x^\top \hat \B  - \frac 1 2   \, \mathrm{diag} \big(  \hat{\mathbf{M}}^\top  \hat \B  \big)    + \ln \bom \pi , \label{eq:koe}
\end{align} 
where $\ln \bom \pi = (\ln \pi_1,\ldots, \ln \pi_G)$,  
$\hat{\mathbf{M}}  = \bmat \muh_1 & \ldots & \muh_G \emat$
and the \emph{coefficient matrix} $\hat \B$ is  defined as, 
\beq \label{eq:B}
\hat \B = \bmat  \hat \beb_1  &  \cdots  &  \hat \beb_G \emat = \hat  \Sig^{-1} \hat{\mathbf{M}}.
\eeq  
The discriminant function in \eqref{eq:koe} is linear in $\x$ with coefficient matrix $\hat \B \in \R^{\pdim \times G}$. Hence, if the $i$-th row of the coefficient matrix  $\hat \B$ is a zero vector $\bo 0$, then the $i$-th  feature does not contribute to the classification rule and hence can be eliminated. If the coefficient matrix $\hat \B$ is row-sparse, then the discriminant rule achieves simultaneous feature elimination.  The  compressive RDA discriminant function is then defined as
\begin{align} \label{eq:koe2}
\hat{\bo d}(\x) &= \big(\hat d_1(\x), \ldots, \hat d_G(\x) \big)  \notag  \\ &=  \x^\top  \Bh  - \frac 1 2   \, \mathrm{diag} \big(  \hat{\mathbf{M}}^\top  \Bh  \big)    + \ln \bom \pi, 
\end{align} 
where  
\beq \label{eq:Bh_apu}
\Bh =  \mathcal H(   \hat \B) 
\eeq 
is a sparsifying transformation of $ \hat \B$. CRDA framework thus requires specifying the  sparsifying transformation $\mathcal H:  \R^{p \times G} \to  \R^{p \times G} $ and the covariance matrix estimator  $\hat  \Sig$.

\begin{figure*}[hbt]
    \centering
    \includegraphics[width=0.8\linewidth]{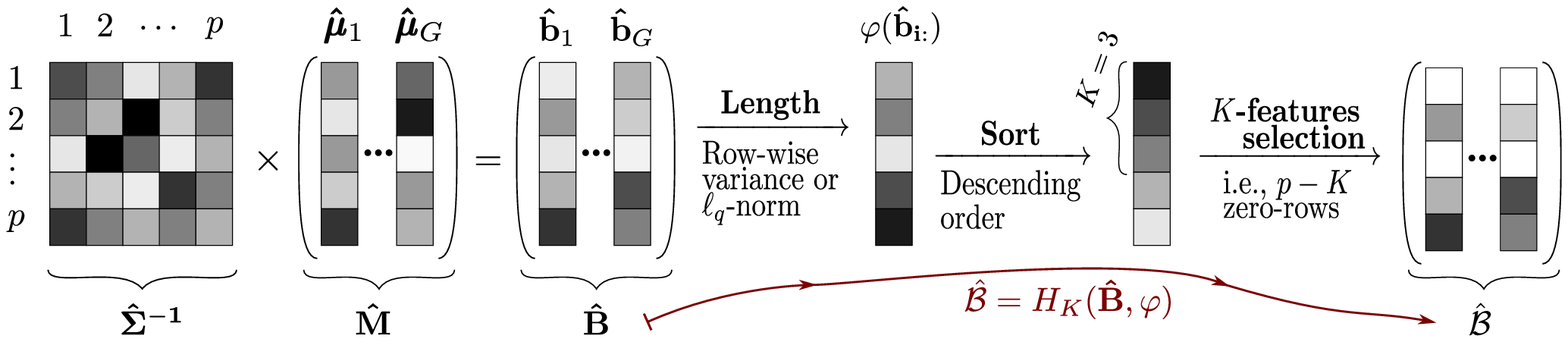}
    \caption{Computing the $K$-rowsparse coefficient matrix $\Bh$ in \eqref{eq:Bh} }
    \label{fig:norm2varsel}
\end{figure*}

The popular Shrunken Centroids Regularized Discriminant Analysis (SCRDA)  \cite{guo2006regularized}   is then an example of CRDA method. It 
uses the element-wise soft-shrinkage operator  $\mathcal S_{\Delta}(\cdot)$ as the sparsifying transform $\mathcal H(\cdot)$. In other words, it computes \eqref{eq:Bh_apu} as   
\beq \label{eq:SCRDA} 
\Bh =  \mathcal S_{\Delta}( \hat \B), 
\eeq 
where the soft-thresholding function is applied element-wise to $\hat \B$. Furthermore, $\Sigt =  \alpha \S + (1- \alpha) \bo I$, is chosen as the covariance matrix estimator. 
A disadvantage of  this approach is that the parameter  $\Delta \in [0, \infty)$ is the same across all groups, and different populations would often benefit from different shrinkage intensity. Another drawback is that an upper bound for shrinkage threshold $\Delta$ is data dependent.   Two-dimensional CV search is  then used  to find the tuning parameter pair $\alpha$ and $\Delta$. 

We promote using different sparsifying transform $\mathcal H(\cdot)$.  Namely, a row-sparse  $\Bh$ can be obtained by using hard-thresholding  (HT) operator:
\beq \label{eq:Bh}
\Bh =  H_K(   \hat \B, \varphi).
\eeq 
Above $H_K(\hat \B, \varphi )$   is defined as a transform that retains only the elements of $K$ rows of $\hat \B$ for which HT-selector function $\varphi : \R^G \to [0,\infty)$ 
produces largest values and set the elements of other rows to zero.   See \autoref{fig:norm2varsel} for an illustration. 
Such CRDA classifier thus uses only $K$ features and is able to do feature selection simultaneously with classification. The tuning parameter $K$ then determines how many features are selected. As HT-selector function $\varphi$, one may use $\ell_q$-norm, 
\[
\varphi_q(\bo b)=\| \bo b \|_q,  \quad \mbox{for $q \in \{1,2,\infty\}$}, 
\]
or the sample variance, 
\[
\varphi_0(\bo b) = s^2(\bo b) = \frac{1}{G-1} \sum_{i=1}^G | b_i - \bar b |^2,
\]
where $\bar b = (1/G)  \sum_{i=1}^G b_i$. Two dimensional CV search described in \autoref{sec:chooseK} can then be used for choosing an appropriate level of joint-sparsity $K$ and the best HT-selector function. 

\subsection{Cross-validation scheme} \label{sec:chooseK}

The joint-sparsity order $K \in \{1, 2, \ldots ,\pdim\}$ and the  HT-selector function  $\varphi \in [\varphi]=\{\varphi_0, \varphi_1, \varphi_2, \varphi_\infty\}$ are chosen using  a cross-validation (CV) scheme. For $K$ we use a logarithmically spaced grid of $K$-values, 
where the upper bound for largest element in the grid is defined as  follows. 
For each $\varphi$, we calculate 
\beq \label{eq:Kub} 
K_{\texttt{UB}}(\varphi) =   \sum_{i=1}^{p} \mathrm{I}(\varphi(  \hat \b_{i:} ) \geq \mathrm{ave}_j \{ \varphi(  \hat \b_{j:} )\} ), 
\eeq 
where  $\hat \b_{i:}$ denotes the $i$-th row  vector of $\hat \B$. 
Thus $ K_{\texttt{UB}}(\varphi_1)$  counts the number of row vectors of $ \hat \B$ that has  $\ell_1$-norm  
larger than the average of $\ell_1$-norms of row vectors of $ \hat \B$. 
As final upper bound $K_{\texttt{UB}}$ we use the minimum value, $K_{\texttt{UB}} = \min_{\varphi \in [\varphi]} K_{\texttt{UB}}(\varphi) $. 
We then use a logarithmically spaced grid of $J= 10$ values $[K]=\{ K_1, K_2, \ldots, K_J\}$, with  $1 <K_1 < K_2 < \cdots < K_J =K_{\texttt{UB}}$  starting from $K_1 = \lfloor 0.05 \cdot p \rfloor$.  
Note that our choice for $ K_{\texttt{UB}}$  adapts to the distribution of $\varphi(  \hat \b_{i:})$. For example,  choosing $ K_{\texttt{UB}} <   \lfloor p \cdot \frac 1 2 \rfloor$ is  recommended when the distribution of $\{\varphi(  \hat \b_{i:} )\}_{i=1}^p$  is such that majority of  $\varphi(  \hat \b_{i:} )$ values are small, i.e., the empirical distribution  of $\varphi(  \hat \b_{i:})$ has a shape that is similar to an exponential distribution.  In such cases, it  is advisable to choose a value for $ K_{\texttt{UB}}$ that is smaller than $  \lfloor p \cdot \frac 1 2 \rfloor$.  This would happen with  our choice of $K_{\texttt{UB}}$ with high probability since the mean value of the exponential distribution is always larger than its median value.

Apart from the way  the lower and upper bounds for the grid of $K$ values are chosen, the used CV scheme is conventional. We split the training data set $\X$ into $Q$ folds and use $Q-1$ folds as a training set and one fold as the validation set. We fit the model (i.e., the CRDA classifier as defined by \eqref{eq:koe2} and \eqref{eq:Bh}) on the training set using $\varphi \in [\varphi]$ and $K \in [K]$ and compute the misclassification error rate on the validation set.  This procedure is repeated so that each of the Q folds is left out in turn as a validation set, and misclassification rates are accumulated.  We then choose the pair in the grid $[\varphi] \times [K]$  that yielded smallest misclassification error rate. In case of ties, we choose the one having smaller value of $K$, i.e., we prefer a model with smaller number of features.

\subsection{CRDA variants}  \label{sec:CRDAclassifiers}

We consider  the following three  CRDA variants:
\begin{itemize} 
\item {\bf CRDA1} uses Ell1-RSCM estimator to compute the coefficient matrix $\hat \B$. The tuning parameters, $K$ and  $\varphi$, are  chosen using the CV procedure of \autoref{sec:chooseK}. 
\item {\bf CRDA2} is as CRDA1 except that $\hat \B$ is computed using the Ell2-RSCM estimator.  
\item {\bf  CRDA3}  uses Rie-PSCM to compute the coefficient matrix $\hat \B$, $\varphi_\infty (\cdot)= \| \cdot \|_\infty$ is used as the HT-selector 
and  $K=K_{\texttt{UB}}(\varphi_\infty)$ as the joint-sparsity level.  
\end{itemize} 
Uniform priors ($\pi_g = 1/G$, $g=1.\ldots,G$) are used in all analysis results. 

The main computational complexity of CRDA methods are related to computing  the chosen covariance matrix estimator $\hat  \Sig$ and the matrix inversion of $ \hat  \Sig$. 
First, we note that the  SVD-trick that is described  in  Appendix~\ref{app:compute_siginv}
reduce the complexity of matrix inversion from  $\mathcal{O}(\pdim^3)$ to $\mathcal{O}(\pdim \ndim^2)$ which is a significant saving in $p \gg n$ case. The complexity of computing the RSCM-Ell1  and RSCM-Ell2 estimators is of the same order as computing the SCM $\bo S$, which  is of order $\mathcal{O}(n \pdim^2)$.  Hence he complexity of CRDA methods are  equivalent to  complexity of computing the underlying covariance matrix estimator in high-dimensional problems.

\subsection{Analysis of real data sets} 

Next, we explore the performance of  CRDA variants in classification task of three different types of data: an image data  (data set \#1),  speech  data (data set \#2) and gene expression  data (data set \#3)  described in \autoref{table:summary}.   The first data set is constructed for classification of five English vowel letters spoken twice by thirty different subjects while the second data set consists of modified observations from the MNIST data set of handwritten digits (namely, digits 4 and 9) and used in NIPS 2003 feature selection challenge. These data sets are available at UC Irvine machine learning repository \cite{Dua:2019} under the names, Isolette and Gisette data set, respectively.   The 3rd data set is from   Khan {\it et al.} \cite{khan2001}  and consists of  small round blue cell tumors  (SRBCT-s) found in children,  and are classified as BL (Burkitt lymphoma), EWS (Ewing's sarcoma), NB (neuroblastoma) and RMS (rhabdomyosarcoma).  The available data is split randomly into training and test sets of fixed length, $n$ and $n_t$, respectively, such that relative proportions of observations in the classes are preserved as accurately as possible.  Then  $L=10$ such random splits 
are performed and averages of error rates are reported. 

The basic metrics used are the test error rate (TER),  feature selection rate (FSR)  and the average computational time (ACT) in seconds. TER is the misclassification rate calculated by classifying the observations from the test set using the classification rule estimated from the training set. FSR is defined as { $\text{FSR} = S/p$, where $S$ denotes the number of features used by the classifier,  and it is a measure of interpretability of the chosen model.  The FSR of CRDA3 will naturally always be higher than FSR of CRDA1 or CRDA2.  

The results of the study reported in \autoref{tab:siginvcomp} is divided in three groups.
The first group consists of the proposed CRDA classification methods, the middle group   consists of  five state-of-the-art classifiers described in \autoref{tab:meth}. The third group consists of three  standard off-the-shelf classifiers commonly used in machine learning, but which do not  perform feature selection. The methods in the third group are 
{\it RF}  which refers to random forest using decision trees (where maximum number of splits in tree was 9 and number of learning cycles was 150), {\it LogitBoost} which refers to LogitBoost method \cite{friedman2000additive}  using stumps (i.e., a two terminal node decision tree) as base learners and 120 as the number of boosting iterations,  and {\it LinSVM} which refers to a linear support vector machine (SVM) using one versus one (OVO) coding design and linear kernel.   Note that since $p \gg n$,  one can always  find a separating hyperplane and hence a linear kernel suffices for SVM.

We first discuss the results of CRDA variants when compared against one another. In the case of data set  \#2, CRDA1 and CRDA3 obtained notably better misclassification rates than CRDA2.  In the case of data set \#3, CRDA1 and CRDA2 obtained perfect 0\% misclassification results while using only 5\% out of all $p=2308$ features. This is a remarkable achievement, especially when compared to TER/FSR rates of the competing estimators. In the case of data set \#1, CRDA3 had better performance than CRDA1 and CRDA2 but it attained this result using more features than CRDA1 and CRDA2. 

Next, we compare CRDA against the methods in the middle group.  
  We notice from \autoref{tab:siginvcomp}   that one of the  CRDA variants is always the best performing method in terms of both accuracy (TER rates) and feature selection rates (FSR).  Only in the case of data set \#2, SCRDA is able to achieve equally good error rate (6.3\%) as CRDA3. 

Finally, we compare the CRDA methods against the methods in the third group.  First we notice that the linear SVM is the best performing  method among the third group, and  for data set \#2  it achieves the best error rate among all methods. However, it chooses {\it all} features. 

As noted earlier, for data set \#3 (Khan {\it et al.}), CRDA1 and CRDA2  attained perfect classification (0\% error rate) while choosing only  5\% out of 2308 genes.  The heat map of {\it randomly selected} 115 genes is shown in \autoref{fig:heatmap}a while  \autoref{fig:heatmap}b shows the heat map of  115 genes picked by CRDA1.  While the former shows random patterns, clearly visible patterns can be seen in the latter heat map. Besides for prognostic classification, the set of  genes found important by CRDA1 can be used  in developing understanding of genes that influence SRBCT-s.

\begin{figure}[!hbt]
\centering
\subfloat[]{\includegraphics[width=0.491\linewidth]{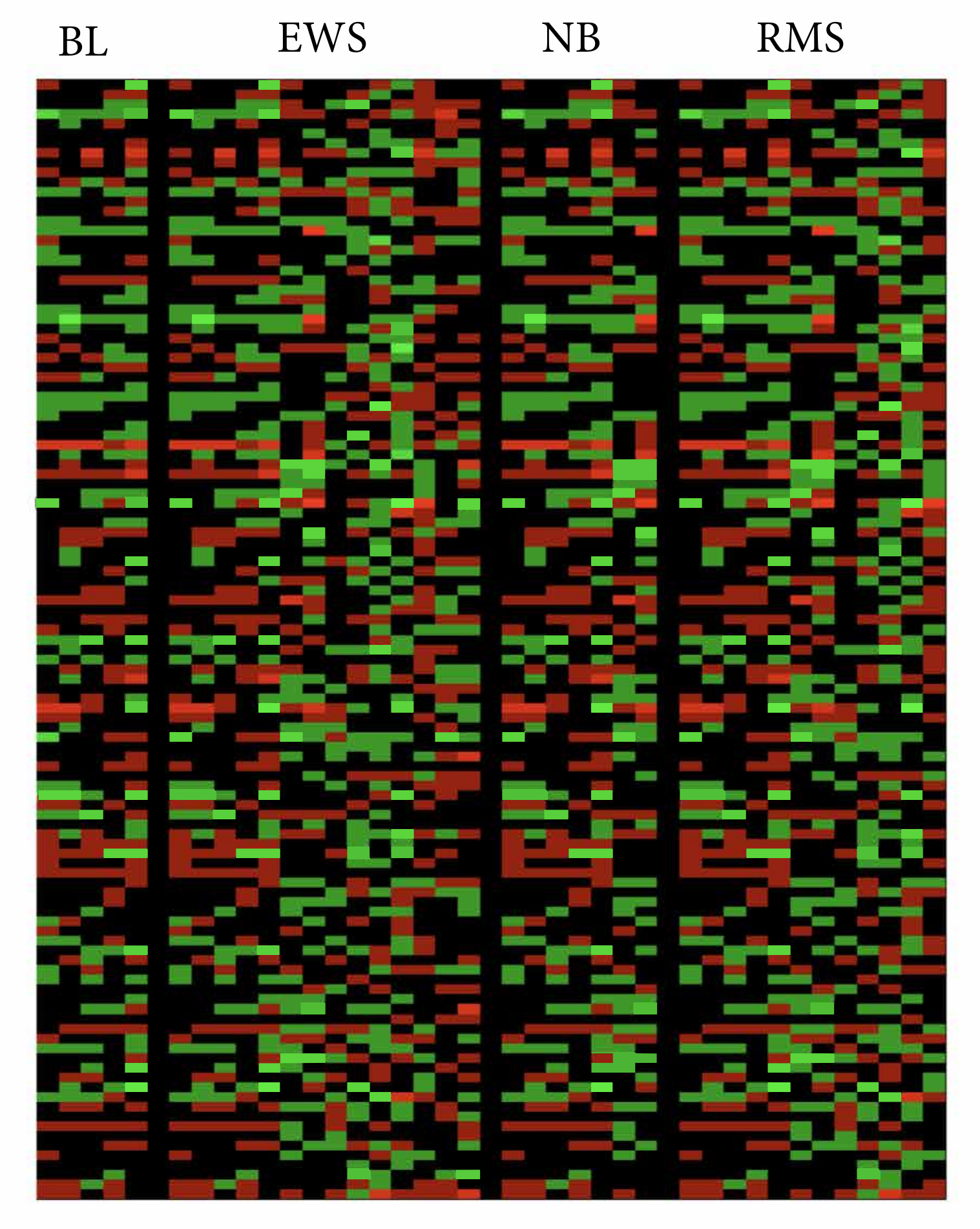}}\hspace{1pt} 
\subfloat[]{\includegraphics[width=0.491\linewidth]{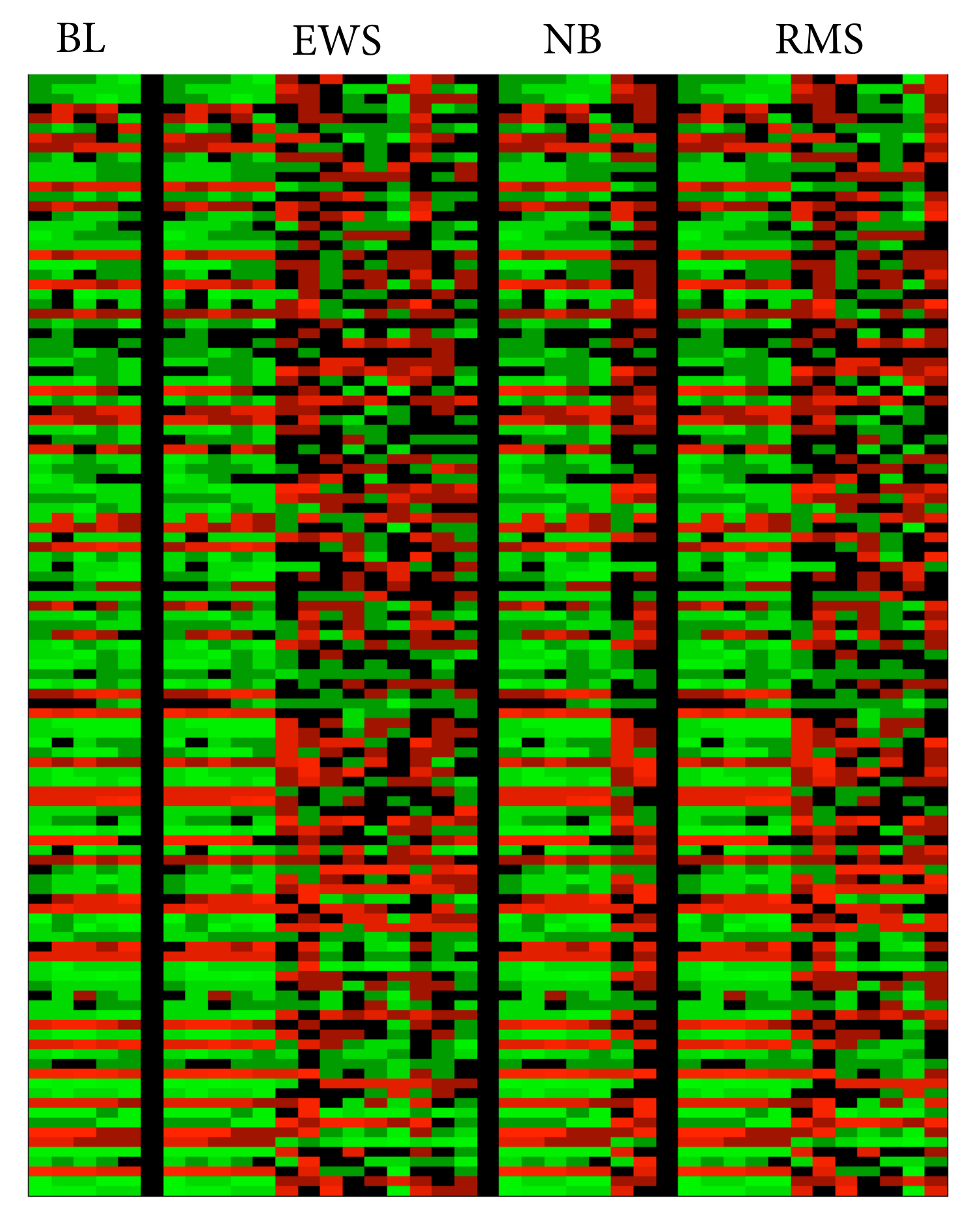}}
\caption{For the data set \#3 (Khan {\it et al.}), the figure displays the heat map of (a) randomly selected 115 genes and (b)  115 genes that were selected by CRDA1 classifier. We note that the CRDA1 classifier obtained  0\% error rate on all 10 Monte Carlo splits of the data to training and test set, and  outperformed all other classifiers.}    \label{fig:heatmap}
\end{figure}

\begin{table}
\caption{Classification results of CRDA variants and their competitors (see \autoref{tab:meth}) for data sets \#1-\#3  in terms of test  error rate (TER) and feature selection rates (FSR) reported in percentages. } \label{tab:siginvcomp}
\centering
\begin{tabular}{ c | r | r  | r | r | r | r | r | } 
\hhline{~-------} 
& \multicolumn{2}{c|}{\#1 Isolet-vowels} &\multicolumn{2}{c|}{\#2 Gisette}   &\multicolumn{2}{c|}{\#3 Khan {\it et al.}}  \\
\hhline{~-------} 
		& TER 	& FSR 	& TER 	& FSR 	& TER 		& FSR      \\ \toprule
CRDA1 	& 2.8		& 21.8	& 7.0		& 18.8 	& {\bf 0} 		& {\bf 5.0}	\\ \hline 
CRDA2 	& 2.7		&{\bf 16.3}	&14.4	& {\bf 8.2} & {\bf 0}		& {\bf 5.0}	\\ \hline
CRDA3	& {\bf 1.3} & 45.1  	& 6.3		& 36.7  	& 1.2  		& 35.5 	\\ \midrule
PLDA 	& 3.6 	&  100   	& 40.5	& 50.8 	& 18.8  		& 82.0    	\\ \hline
SCRDA 	& 1.6 	&  88.2   	& 6.3 	& 87.0 	& 2.8  		& 81.5 	\\ \hline
varSelRF 	& 3.0 	&  28.9   	& 7.9		& 18.8	& 2.4  		& 78.8   	\\ \hline
Logit-ASPLS 	& 26.4 	&  98.6   	& 27.8	& 90.0 	& 43.2  	& 100    	\\ \hline
SPCALDA 	& 19.1 	&  73.7   	& 6.4 	& 47.4 	& 3.2    	& 46.9    	\\  \midrule
SVM		& {\bf 1.3} & 100.0 	& {\bf 6.0}	& 100.0	& 0.8			& 100.0 	\\ \hline
LogitBoost& 3.0	& 100.0	& 7.7		 & 100.0	& 10.4 		& 100.0       \\ \hline 
RF	         & 1.8 	& 100.0	& 9.5		&  100.0	& 1.2   		& 100.0        \\ \bottomrule
\end{tabular}
\end{table}

\begin{table*}
\caption{Summary of the real data sets used in this paper.} 	\label{table:summary}
\centering
\begin{tabular}{l | c | c |  c |   c  | c | c} 
\toprule
data set 									&$p$ 	&$n$ 	&$n_t$	&$\{n_1,\ldots,n_G\}$ 	&$G$	&Description \\ 
\midrule
 \#1 Isolet-vowels  \cite{Dua:2019}				&617		&100		&200		&\{20,20,20,20,20\}		&5		&Spoken English vowel letters \\
 \#2 Gisette  	\cite{Dua:2019}					&5000	&350		&6650	&\{175,175\}			&2		&OCR digit recognition ('4' vs '9') \\
 \#3 Khan {\it et al.}  	\cite{khan2001}				&2308	&38		&25		&\{5, 14, 7, 12\}			&4		&SRBCTs \\ 
 \#4 Su {\it et al.}  \cite{Su:2002cd}				&5565	&62		&40		&\{15, 16,  17,  14\}		&4		&Multiple mammalian tissues \\
 \#5 Gordon {\it et al.}  \cite{Gordon2002} 			&12533	&109		&72	 	&\{90, 19\}				&2		&Lung Cancer \\
 \#6 Burczynski  {\it et al.} \cite{Burczynski2006}	&22283	&76		&51		&\{35,   25,    16\}		&3		&Inflammatory bowel diseases (IBD) \\
 \#7 Yeoh {\it et al.}  \cite{yeoh2002} 				&12625	&148		&100		&\{9,16,  38, 12, 26, 47\}	&6 		&Acute lymphoblastic leukemia (ALL) \\
 \#8 Tirosh  {\it et al.}   \cite{tirosh2006}			&13056		&114 	&77		&\{23, 22,  23,  46\}		&4		&Yeast species \\
 \#9 Ramaswamy{\it et al.} \cite{ramaswamy2001} 	&16063	&117		&73		&\{7 ,6, 7, 7,13, 7, 6, 6, 18, 7, 7, 7, 7,12\}  &14 	&Cancer \\
\bottomrule
		\end{tabular}
\end{table*}

\section{Feature selection performance}   \label{sec:psd}

Next we create a partially synthetic gene expression data set using the 3rd data set  in \autoref{table:summary} to study how well the different classifiers are able to detect differentially expressed (DE) genes. The data set \#3 consists of  expression levels of $p=2308$ genes and we randomly kept $p_1=115$ (i.e., $\approx 5\%$ of $p$) covariates of gene expression levels as significant features,  i.e., differentially expressed, and the rest $p_0=2193$ irrelevant (non-DE) features are generated from $\mathcal{N}_{n} (\bo 0, 0.01\times\mathbf{I})$. 

Since we now have a ground truth of differentially expressed genes available, we may evaluate the estimated models using the false-positive rate (FPR) and the  false-negative rate (FNR). A false-positive is a feature that is picked by the classifier but is not DE in any of the classes  in the sense that  $\mu_{g,j}=0$ for all $g=1,\ldots,G$.   A false-negative is a feature that is DE in the true model in the sense that $\mu_{g,j} \neq 0$ for all $g=1,\ldots,G$, but is not picked by the classifier. Formally, 
 \[
\text{FPR} =  \frac{F}{p_0} \qquad \text{and} \qquad  \text{FNR} = \frac{p_{1}  -  T}{p_1} , 
\]
where $p_1$ is the number of truly positive (DE) features, $p_{0}$ is the number of truly negative (non-DE) features, $F$ is the number of false positives and $T$ is the number of true positives, i.e., $S=F+T$ gives the total number of features picked by the classifier. Again, recall that the sparsity level $K$ used in the hard-thresholding operator is the number $S$ in the case of CRDA methods.  Note that both FPR and FNR should be as small as possible for exact and accurate selection (and elimination) of genes.  

First, we compare the gene selection performance of CRDA with the conventional gene selection approach \cite{cui2003statistical} which performs a t-test and computes the p-values using permutation analysis with 10,000 permutations.  A gene is considered to be differentially expressed when it shows both statistical and biological significance. Statistical significance is attested when the obtained p-value of the test statistics is below the  significance level (false alarm rate), say $p_{\mathrm{FA}}=0.05$. Biological significance is attested when the log ratio of fold change  of mean gene expressions  is outside the interval  $[-\log_2(c), \log_2(c)]$ for some cutoff limit $c$, say $c=1.2$. \autoref{fig:genesel}(a) display such a  volcano plot which displays the -$\log_{10}$ of p-values  of $t$-tests for each genes against the $\log_2$  fold change. The used statistical significance level is shown as the dotted horizontal line and the biological significance bounds as dotted vertical lines. 

The plot illustrates that there are 19 out of 115 differentially expressed genes which are both statistically and biologically significant  (genes on the upper left and upper right corners of the plot) along with a few which are biologically significant but not statistically significant (lower left and lower right corners).  When considering each class in turn against the rest, then 60 out of a total of 115 DE genes are picked as being DE. The results for CRDA2 based gene selection is shown in \autoref{fig:genesel}(b) for the same data set where we have plotted the gene index, $i$, against the sample variance, $s_i$,  of the $i$th row vector of the coefficient matrix $\hat{\bo B}$. From 115 DE genes, CRDA2  has selected 107 DE genes.

\begin{figure}[!htb]
 \centering
\subfloat[]{\includegraphics[width=0.48\linewidth]{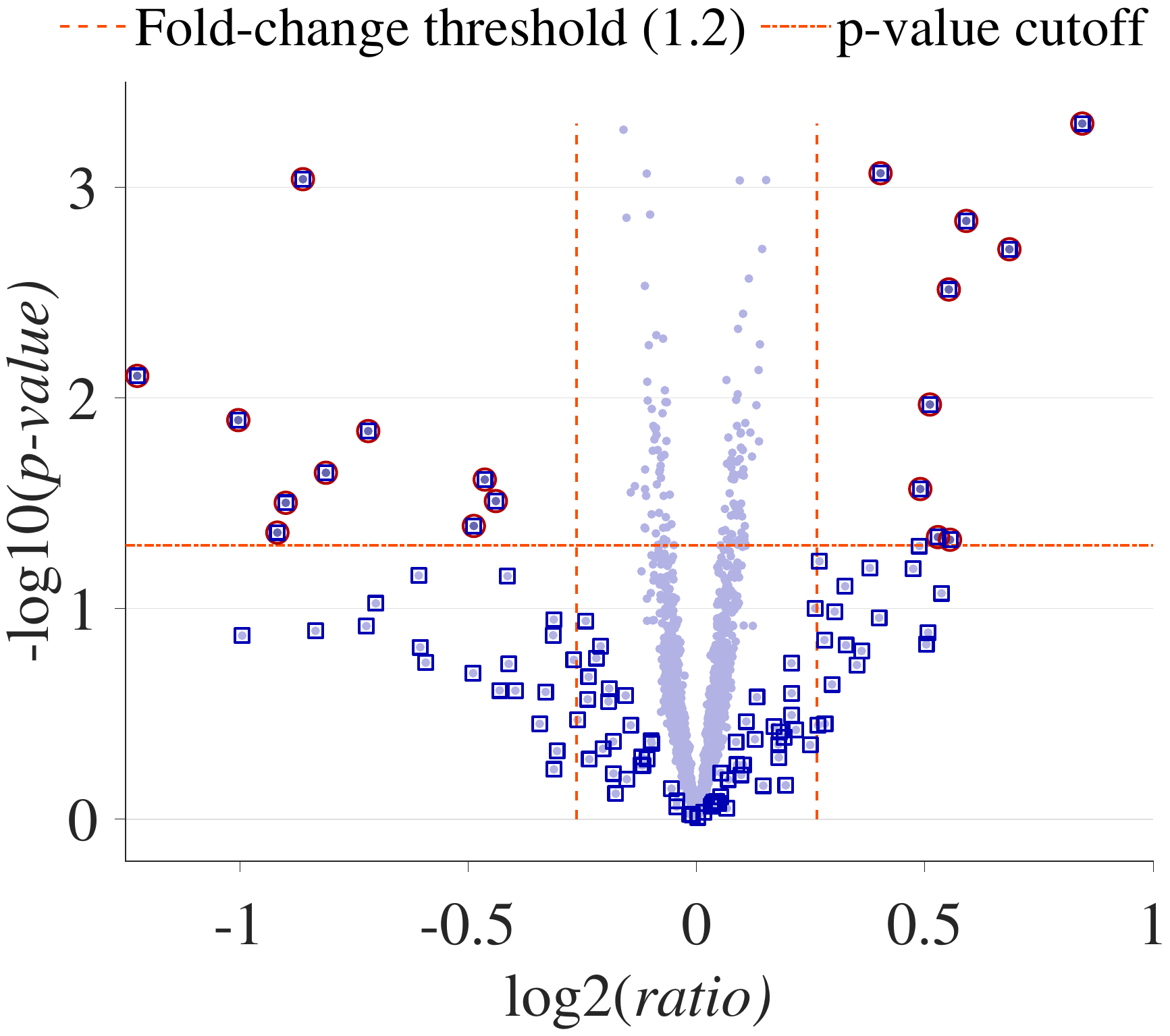}} \:
\subfloat[]{\includegraphics[width=0.49\linewidth]{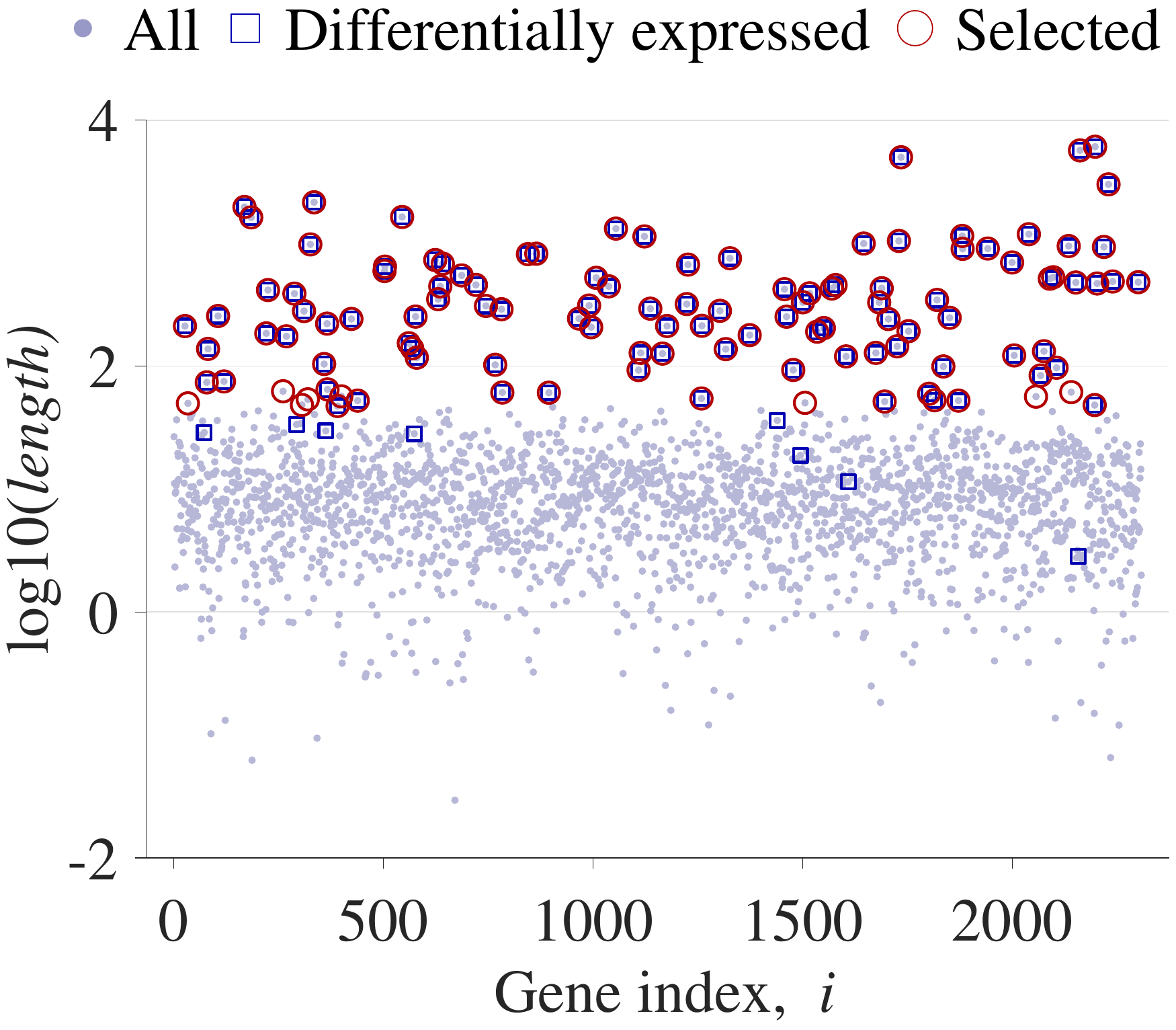}}
\captionof{figure}{Feature selection results for the partially synthetic genomic data. 
(a) Volcano plot illustrating that  19 (9 up-regulated and 10 down-regulated) genes are selected when comparing the third class against the others. 
When each class are compared in turn against the others,  then 60 out of 115 DE genes are picked. 
(b)  Plot of gene index, $i$, versus the  $\log_{10}$ of the length of the $i$th row vector of the coefficient matrix $\hat{\bo B}$ produced by HT-selector  $\varphi_0$, i.e., $length = \varphi_0(\hat \b_{i:})$.  CRDA2 has selected 107 differentially expressed genes out of total 115.}
    \label{fig:genesel}
\end{figure}

\begin{table*}
\caption{State-of-the-art methods included in our comparative study. All of the included methods are designed with an aim of performing feature selection during the learning process. Tuning parameters are chosen using the default method implemented in the R-packages, including the grids for parameter tuning.   When CV is used, then a five-fold CV is selected. } 	\label{tab:meth}
\centering
\begin{tabular}{l  | l  | l | r} 
		\toprule
Method		& Reference 															& R-package or other		& Version\\ \midrule
PLDA		& Penalized LDA \cite{witten2011plda}  										& {\tt penalizedLDA} & 1.1 \\ 
SCRDA		& Shrunken centroids RDA  \cite{guo2006regularized}  							& {\tt rda} 			& 1.0.2.2 \\ 
varSelRF		& Random forest (RF) with variable selection  \cite{varSelRF} 						& {\tt varSelRF} 	& 0.7.8 \\  
Logit-ASPLS 	& Adaptive sparse partial least squares based logistic regression \cite{logitASPLS2018}  	& {\tt plsgenomics} 	& 1.5.2 \\ 
SPCALDA		& Supervised principal component analysis based LDA  \cite{SPCALDA2018} 			& {\tt SPCALDA} 	& 1.0 \\ 
Lasso 		& Regularized multinomial regression with $ \ell_1$ (Lasso) penalty  \cite{friedman2010regularization}	& {\tt GLMNet} 	& 3.0-2 \\
EN			& Regularized multinomial regression with Elastic Net (EN) penalty  \cite{friedman2010regularization}	& {\tt GLMNet} 		& 3.0-2 \\
\bottomrule
\end{tabular}
\end{table*}

As a baseline, we further compute the test error rate of the {\sl naive classifier} that assigns all observations in the test set to the class that has the most observations in the training set. \autoref{fig:ter_sds}(a) displays the box plots of FSR and TER of CRDA methods and three state-of-the-art classifiers. Again the CRDA classifiers obtained the smallest average TER  while choosing  the smallest number of features. \autoref{fig:ter_sds}(b) provides the box plots of FPR and FNR for all methods. The proposed CRDA methods offer the best feature selection accuracy. 
 On the contrary, SCRDA and varSelRF  methods produce large false-positive rates, meaning that they have chosen a large set of genes as DE while they are not. PLDA is performing equally well with CRDA in terms of FPR/FNR rates, but its misclassification rates are considerably larger. Furthermore, CRDA methods exhibit the smallest amount of variation (i.e., standard error) in all measures (TER/FSR/FPR/FNR). We also notice that varSelRF does not perform very well compared to CRDA or other regularized LDA techniques like SCRDA. 

\begin{figure}[!hbt]
 \centering
 \subfloat[]{\includegraphics[width=0.98\columnwidth]{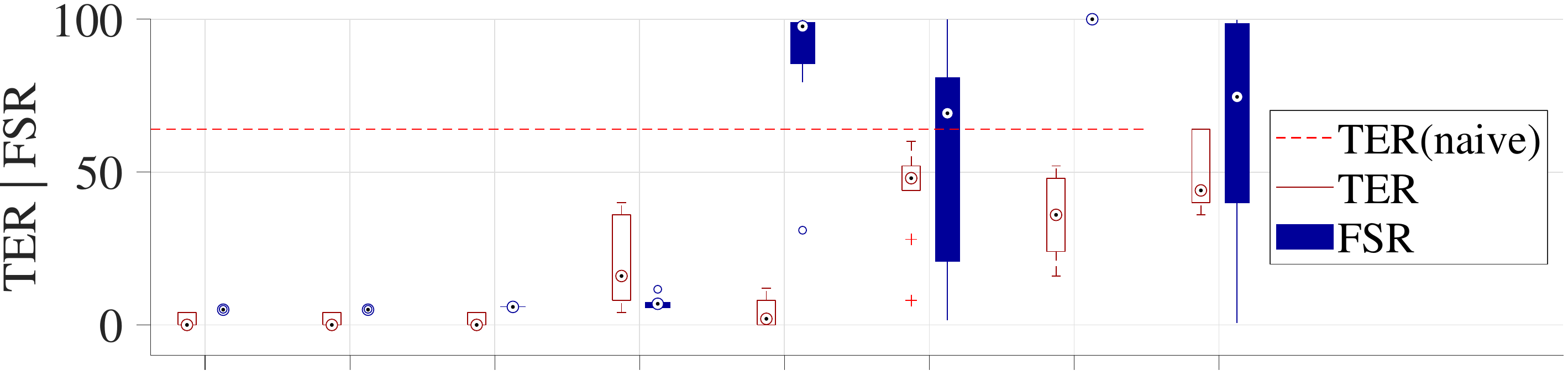}} \hspace{2pt}
\subfloat[]{ \includegraphics[width=\columnwidth]{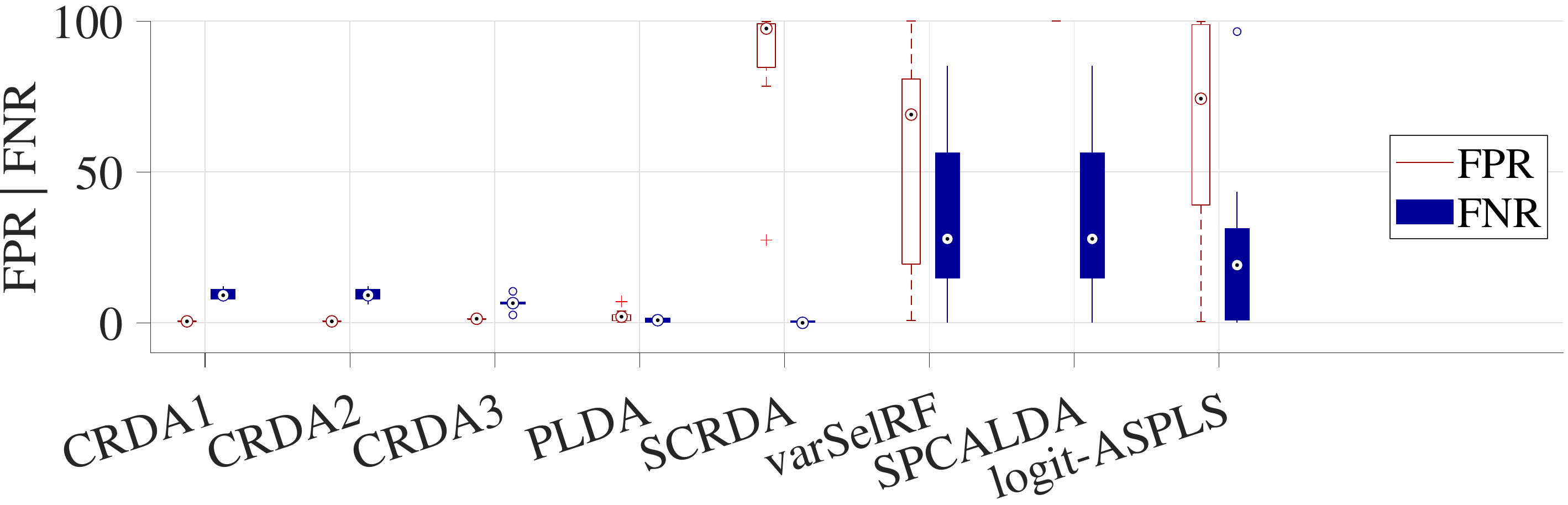}}
  \caption{Classication accuracy results for the partially synthetic gene expression data. (a) Boxplots of  test error rates (TER) and feature selection rates (FSR). (b) Boxplots of  false negative rates (FNR) and false positive rates (FPR) reported in percentages. Results are from simulation study of 10 random splits of the data to training set and test set.}
    \label{fig:ter_sds}
\end{figure}

\section{Analysis of real gene expression data} \label{sec:results}

We compare the performance of the proposed CRDA methods to seven state-of-the-art classifiers on a diverse set of real gene expression (microarray) data sets.  These data sets  \#4 -- \#9 in \autoref{table:summary} constitute a representative set of  gene expression data sets available in the open literature, having a varying number of features ($p$), sample sizes ($n_g$) or classes ($G$).  As earlier, in each Monte-Carlo trial, the data is divided randomly into training sets and test sets of fixed length, $n$ and $n_t$, respectively, such that relative proportions of observations in the classes are preserved as accurately as possible. We do not include CRDA3 in this study due to incurred high computational cost related to computing the  Rie-PSCM covariance matrix estimator. Furthermore, since the performance of LogitBoost and Random Forest (RF) were inferior to LinSVM, we do not report the performance of these methods in the study.

The Gene Expression Omnibus (GEO) data repository contains the data sets \#6 and \#8 with accession numbers GDS1615 and GDS2910, respectively.    The 9th data set (also known as the global cancer map (GCM) data set) is  available online \cite{GCMdata}. 
The rest of the data sets, namely data sets 4, 5, and 7 are included in the R-package {\tt datamicroarray}\footnote{https://github.com/ramhiser/datamicroarray}.

\autoref{fig:ter_rds} displays the average TER and FSR values.  CRDA method obtained the lowest error rates for 5 out of 6 data sets considered; only for data set \#5 (Gordon), Logit-ASPLS  had better performance than CRDA. Furthermore, CRDA achieved the top performance by picking a largely reduced number of genes: the number of genes selected ranged from 5.1\% to 23.4\%. Only Lasso and EN selected  fewer number of genes than CRDA methods but their error rates were much higher in all cases except for data set \#5 (Gordon).

\begin{figure*}[!ht]
    \centering
    \includegraphics[width=\linewidth]{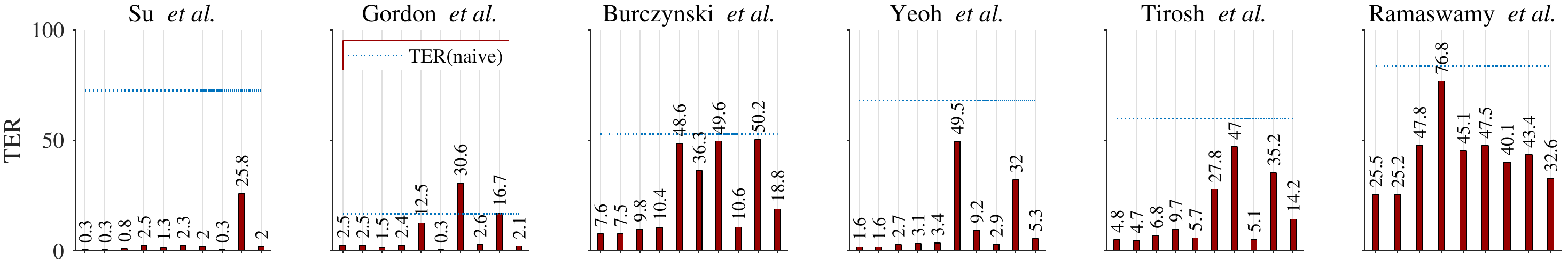} \\ \vspace{5pt}
    \includegraphics[width=\linewidth]{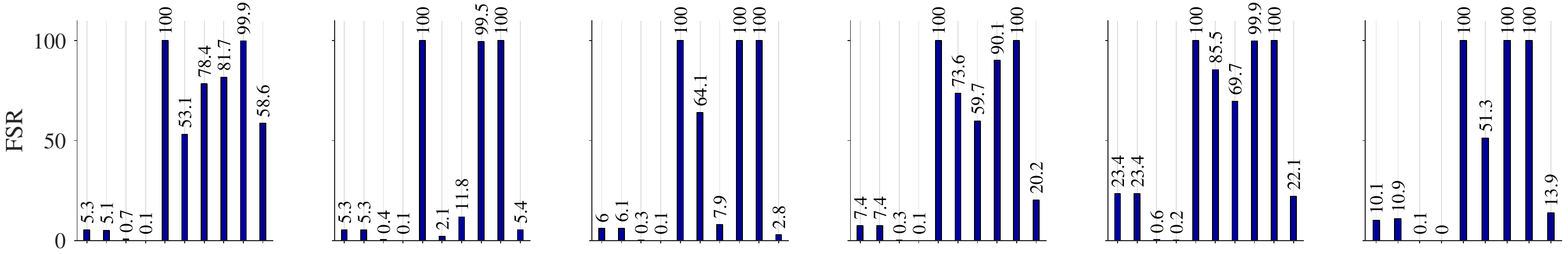} \\ \vspace{5pt}
    \includegraphics[width=\linewidth]{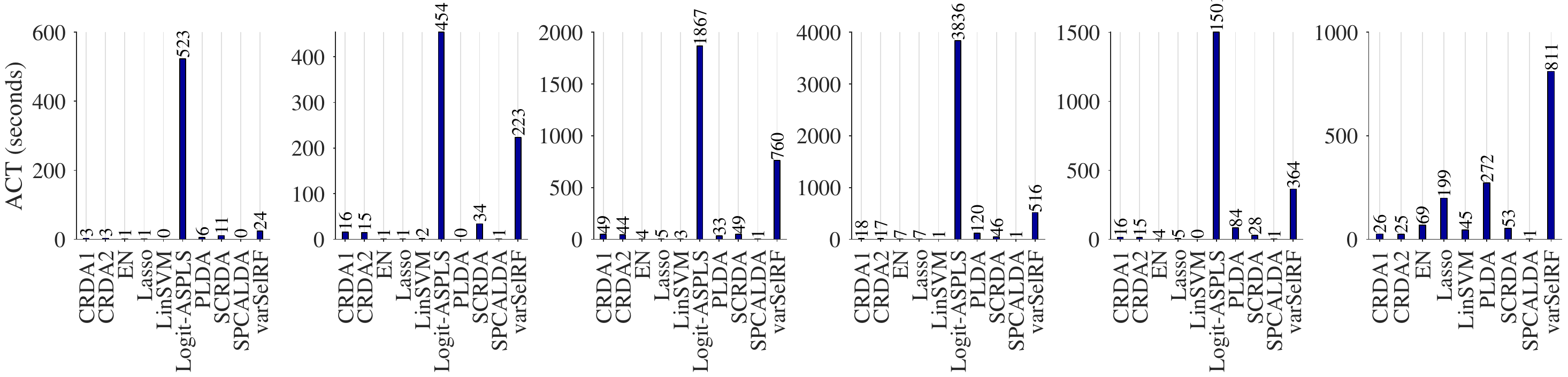}
      \caption{Test error rates (TER) and feature selection rates (FSR) in percentages, and average computational times (ACT) in seconds are reported for the last six real data sets given in \autoref{table:summary}. Results are averages over $L = 10$ random splits of data into training and test sets.} \label{fig:ter_rds}
\end{figure*}

For some data sets, such as for the seventh (Yeoh)  and eight (Tirosh), SCRDA and CRDA obtained comparable TER values, but the former fails to eliminate as many genes.   Again we observe that many of the competing classifiers, such as varSelRF, PLDA or SPCALDA are performing rather poorly for some data sets; this is especially the case for the sixth  (Burczynski) and ninth (Ramaswamy) data sets.   

Data set \#9  (Ramaswamy) presents a very difficult data set for any any classification method: it contains  $G=14$ classes,  $p=16063$ genes and only 117 observations in the training set.   Results obtained by CRDA clearly illustrates its top performance compared to others in very demanding HDLSS data scenarios. CRDA classifiers outperformed most of the methods with a significant margin: CRDA gives smallest error rates (25.2\%) while at the same time reducing the number of features (genes) used in the classification procedure.  The next best method is varSelRF, but it is still far behind (32.6\%).   We note that Logit-ASPLS method failed for this data set due to computational and memory issues and hence it is excluded.

Finally, bottom panel of \autoref{fig:ter_rds} displays the ACT values (in seconds)  of the methods. Variation in ACT between different data sets is due to different number of features ($p$) and observations ($n$) of data sets.  In terms of computational complexity, CRDA looses to Lasso, EN and LinSVM. However, compared to other methods that are designed to perform feature selection, CRDA is among the most efficient method.  Particularly, one can note that Logit-ASPLS and varSelRF are computationally very intensive.

\section{Discussions and Conclusions} \label{sec:concl}

We proposed a novel compressive regularized discriminant analysis (CRDA) method for high-dimensional  data sets, where data dimensionality $p$ is much larger than the sample size  $n$ (i.e., $p \gg n$, where $p$ is very large, often several thousands). 

The method showcase accurate classification results and feature selection ability while being fast to compute in real-time. Feature selection is important for post-selection inference in various problems as it provides insights on which features (e.g., genes) are particularly important in expressing a particular trait (e.g., cancer type).

Our simulation studies and real data analysis examples illustrate that CRDA offers top performance in all three considered facets: classification accuracy, feature selection, and computational complexity. Software (both MATLAB and R)  toolboxes to reproduce the results are available: see \autoref{sec:intro} for links to the packages.

\begin{appendices}

\section{Tuning parameter selection of Ell1- and Ell2-RSCM estimator} \label{app:alpha_est}

Ell1-RSCM and Ell2-RSCM estimators \cite{ollila2019optimal}  computes an estimate  $\hat \alpha$ of the optimal shrinkage parameter $\alpha_o$ in \cite{ollila2019optimal} 
as 
\beq \label{eq:alpha_est}
	\hat \alpha =  \dfrac{( \hat \gamma-1)}{(  \hat \gamma - 1) + \hat \kappa (2  \hat \gamma  + p)/\ndim + ( \hat \gamma  + p)/(n-1) }, 
\eeq 
where $\hat \gamma$ is an estimator of the sphericity parameter $\gamma = p \tr(\Sig^2)/\tr(\Sig)^2$ and $\hat \kappa$ is an estimate of the elliptical kurtosis parameter $\kappa= (1/3) \mathrm{kurt}(x_i)$, calculated as the average sample kurtosis of the marginal variables and scaled by $1/3$.   
The  formula above is derived under the assumption that the data is a random sample from an unspecified elliptically symmetric distribution
with mean vector $\bom \mu$ and covariance matrix $\Sig$. 

Ell1-RSCM  and Ell2-RSCM estimators differ only in the used estimate $\hat \gamma$ of  the sphericity parameter in \eqref{eq:alpha_est}. 
Estimate of sphericity used by Ell1-RSCM is
\begin{align}  \label{eq:estimatorsquared}
	\hat \gamma  =  \min\left(p, \max\left(1 \, , \, \frac{n}{n-1}\left( p \tr(\tilde \S^2) -
	\frac{p}{n}\right) \right) \right)
\end{align}
where $\tilde \S$ is the \emph{sample spatial sign covariance matrix}, defined as
\begin{align*}
	\tilde \S &= \frac{1}{n} \sum_{i=1}^{n} \frac{(\x_{i} - \hat{\boldsymbol{\mu}})(\x_{i} - \hat{\boldsymbol{\mu}})^\top}{\|\x_{i} -\hat{\boldsymbol{\mu}}\|^2},
\end{align*}	
where $\hat{\boldsymbol{\mu}} = \argmin_{\boldsymbol{\mu}} \sum_{i=1}^{n}\|\x_{i} - \boldsymbol{\mu}\|$ is the
spatial median.   In Ell2-RSCM, the estimate of sphericity $\hat \gamma$ is computed as 
\begin{equation}\label{eq:hatgamma_ell2}
\hat \gamma  
	=   \min\left(p, \max\left(1 \, , \,  \hat b_n \Big( \frac{p\tr(\S^2)}{\tr(\S)^2} - \hat a_n \frac{p}{n} \Big) \right) \right) , 
\end{equation}
where 
\begin{align*}
	\hat a_n &=  \left(\frac{n}{n+ \hat \kappa} \right) \left(  \frac{n}{n-1} + \hat \kappa \right)  \\ 
\hat 	b_n &= \frac{ (\hat \kappa  + n)(n-1)^2}{ (n-2)(3  \hat \kappa (n-1) + n(n+1))} 
\end{align*}
and $\bo S$ is the pooled SCM in \eqref{eq:pooledSCM}.

\section{On computing $\Sigit$}  \label{app:compute_siginv}

The matrix inversion can be performed efficiently using the SVD-trick  \cite{guo2006regularized}. The SVD of $\X$ is  
\[
\X= \mathbf{U} \mathbf{D} \mathbf{V}^\top,
\]
where $\mathbf{U}  \in \R^{p \times m}$, $\mathbf{D} \in \R^{m \times m}$, $\mathbf{V} \in \R^{n \times m}$ and $m = \mathrm{rank}(\X)$. 
Direct computation of SVD is time consuming and the trick is that $\mathbf{V}$ and $\mathbf{D}$ can be computed first from SVD of $\X^\top \X = \tilde{\mathbf{V}} \tilde{\mathbf{D}}^2 
\tilde{\mathbf{V}}^\top$,  which is only an $n \times n$ matrix. Here  $\tilde{\mathbf{V}}$ is an orthogonal $ \ndim \times \ndim$  matrix whose first $m$-columns are $\mathbf{V}$ and $\tilde{\mathbf{D}}$ is an $ \ndim \times \ndim$ diagonal matrix whose upper left corner $m \times m$ matrix is $\mathbf{D}$. 
After we compute  $\mathbf{V}$ and $\mathbf{D}$ from SVD of $\X^\top \X$, we may compute $\mathbf{U}$ from $\X$ by $\mathbf{U} = \X \mathbf{V} \mathbf{D}^{-1}$. 
Then, using the SVD representation of the SCM, $\S = (1/n) \mathbf{U}\mathbf{D}^2 \mathbf{U}^\top$,  and simple algebra, one obtains a simple formula for the inverse 
of the regularized SCM in \eqref{eq:rscm} as:
\begin{align*}\label{eq:RSCMinv} 
\Sigit &= \mathbf{U} \bigg[ \Big( \frac{\alpha}{n}\,\mathbf{D}^2 + (1- \alpha)  \eta \bo I_m  \Big)^{-1} - \frac{1}{(1- \alpha)\eta} \bo I_m \bigg] \,\mathbf{U}^\top \\ &+ \frac{1}{(1- \alpha) \eta} \bo I_\pdim,
\end{align*}
where $\eta = \tr(\S)/\pdim = \tr(\mathbf{D}^2)/n\pdim$.
This reduces the complexity from  $\mathcal{O}(\pdim^3)$ to $\mathcal{O}(\pdim \ndim^2)$ which is a significant saving in $p \gg n$ case.

\end{appendices}

\section*{ACKNOWLEDGMENT}

The research was partially supported by the Academy of Finland grant no. 298118 which is gratefully acknowledged.

\bibliographystyle{IEEEtran}


\begin{IEEEbiography}[{\includegraphics[width=1in,height=1.25in,clip,keepaspectratio]{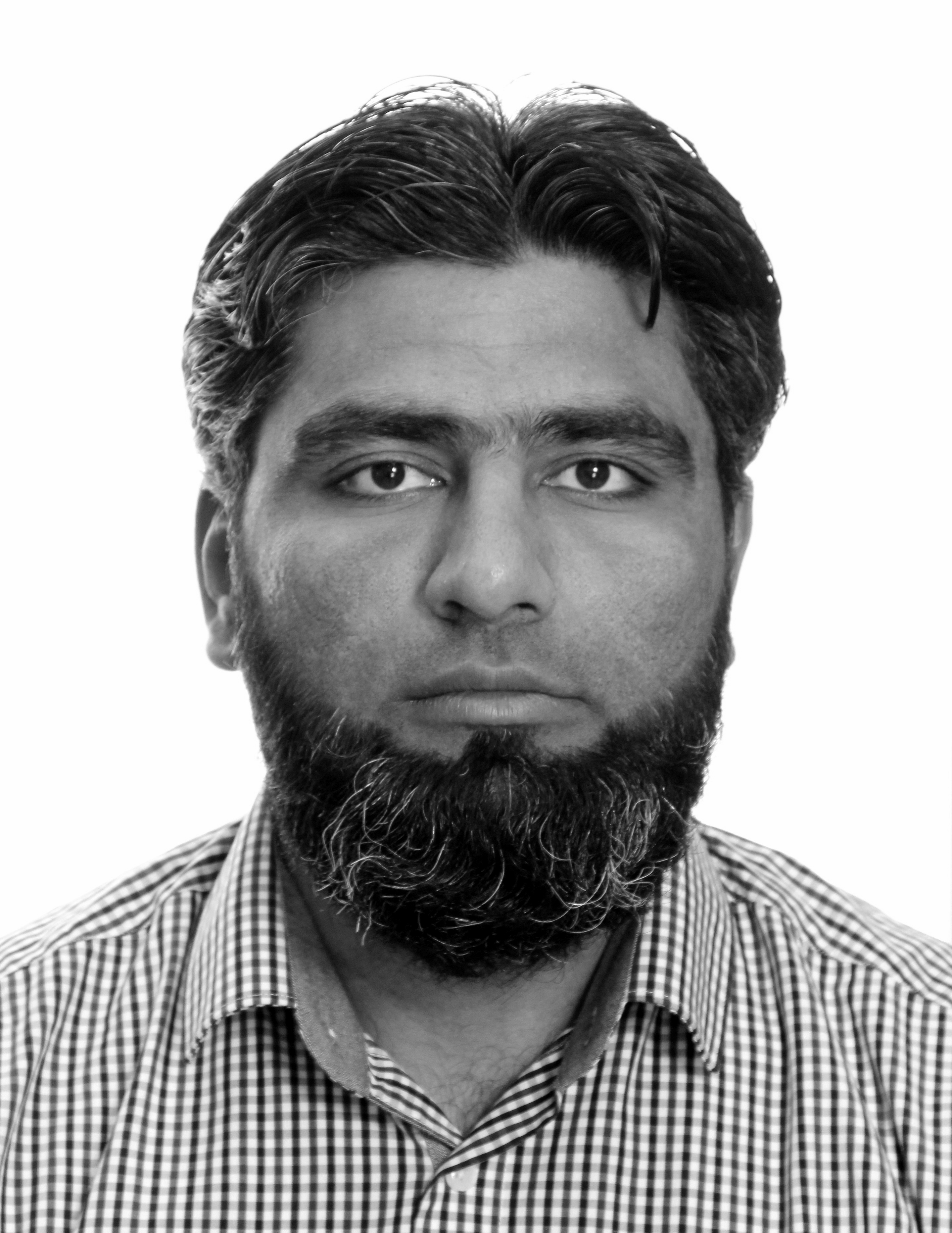}}]{MUHAMMAD NAVEED TABASSUM} (S'14) is currently working as R\&D Engineer at Nokia, Espoo, Finland. He received the D.Sc. (Tech.) degree in signal processing and data science from Aalto University, Espoo, Finland, in 2020, and the M.Sc. degree in Electrical Engineering from King Saud University, Riyadh, Saudi Arabia, in 2015, where he worked in multidisciplinary teams on funded research projects at NPST and RFTONICS research centers. Earlier, he developed iOS apps while working in agile software development teams at Coeus Solutions GmbH, Lahore Branch, Pakistan. His career interests involve research and development of advanced solutions to the signal processing and data science based problems related to natural language processing, Internet of Things, wireless communication and sensor networks, and bioinformatics applications, particularly focusing statistical-, machine- and deep-learning approaches.
\end{IEEEbiography}

\begin{IEEEbiography}[{\includegraphics[width=1in,height=1.25in,clip,keepaspectratio]{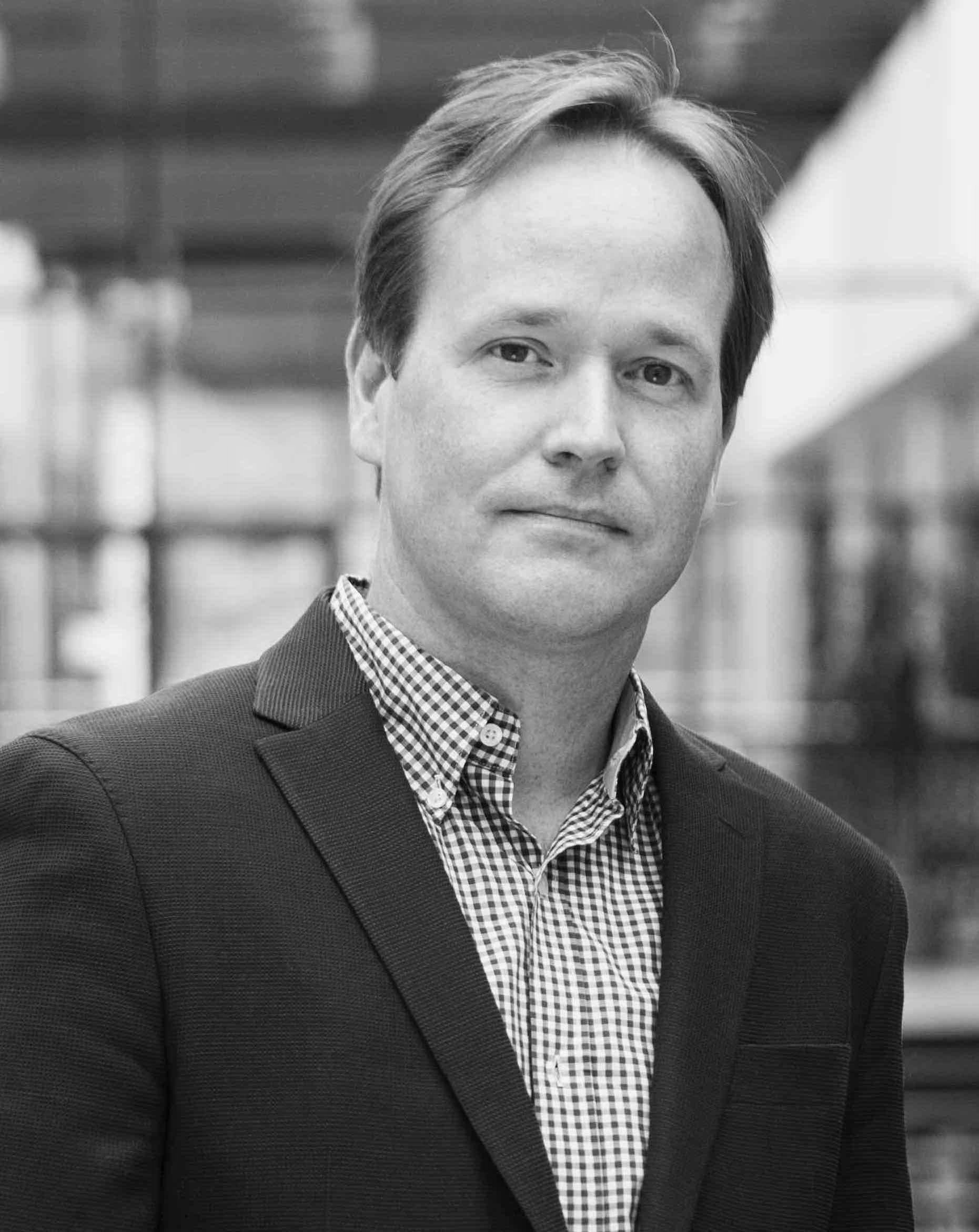}}]{Esa Ollila} (M'03) received the M.Sc. degree in mathematics
from the University of Oulu, Oulu, Finland, in
1998, Ph.D. degree in statistics with honours from the
University of Jyvaskyla, Jyvaskyla, Finland, in 2002,
and the D.Sc.(Tech) degree with honours in signal
processing from Aalto University, Aalto, Finland, in
2010. From 2004 to 2007, he was a Postdoctoral Fellow
and from August 2010 to May 2015 an Academy
Research Fellow of the Academy of Finland. He has
also been a Senior Lecturer with the University of
Oulu. Currently, since June 2015, he has been an Associate
Professor of Signal Processing, Aalto University, Finland. He is also an
Adjunct Professor (Statistics) of Oulu University. During the Fall-term 2001,
he was a Visiting Researcher with the Department of Statistics, Pennsylvania
State University, State College, PA while the academic year 2010--2011 he spent
as a Visiting Postdoctoral Research Associate with the Department of Electrical
Engineering, Princeton University, Princeton, NJ, USA. He is a member of
EURASIP SAT in Theoretical and Methodological Trends in Signal Processing
and co-author of a recent textbook, {\it Robust Statistics for Signal Processing},
published by Cambridge University Press in 2018.  Recently, he organized a 
special session on Robust Statistics in Signal Processing  in CAMSAP 2019 and a special issue 
 on {\it Statistical Signal Processing Solutions and Advances for Data Science}  in 
 Signal Processing (Elsevier). His research interests lie in
the intersection of the fields of statistical signal processing, high-dimensional
statistics, bioinformatics and machine learning with emphasis on robust statistical methods and
modeling.
\end{IEEEbiography}

\end{document}